\documentclass[runningheads]{llncs}
\usepackage[T1]{fontenc}
\usepackage[utf8]{inputenc}
\usepackage{graphicx}
\usepackage{color}
\usepackage{booktabs}
\usepackage{tabularx}
\usepackage{caption}
\usepackage{multirow}
\usepackage{amssymb}
\usepackage{amsmath,amsfonts,amssymb}
\usepackage{hyperref}
\begin{document}
\title{VLM-PAR: A Vision Language Model for Pedestrian Attribute Recognition}

\author{Abdellah Zakaria Sellam\inst{1,2}\orcidID{0009-0003-6876-2220} \and
Salah Eddine Bekhouche\inst{3}\orcidID{0000-0001-5538-7407} \and
Fadi Dornaika\inst{3,4}\orcidID{0000-0001-6581-9680}
\and
Cosimo Distante\inst{1,2}\orcidID{0000-0002-1073-2390}
 \and
Abdenour Hadid\inst{5}\orcidID{0000-0001-9092-735X}
 }
\authorrunning{AZ. Sellam {\it et al.}

}

\institute{
Department of Innovation Engineering; University of Salento, Italy  \\
\email{abdellahzakaria.sellam@unisalento.it}\\
\and
Institute of Applied Sciences and Intelligent Systems - CNR, Lecce, Italy\\
\email{cosimo.distante@cnr.it}\\
\and
University of the Basque Country UPV/EHU, San Sebastian, Spain
\email{sbekhouche001@ikasle.ehu.eus, fadi.dornaika@ehu.eus}\\
\and
IKERBASQUE, Basque Foundation for Science, Bilbao, Spain\\
\and
\email{abdenour.hadid@sorbonne.ae}
\and
 Sorbonne University Abu Dhabi, Abu Dhabi, UAE \\
}

%
\maketitle              
\begin{abstract}
Pedestrian Attribute Recognition (PAR) involves predicting fine‐grained attributes such as clothing color, gender, and accessories from pedestrian imagery, yet is hindered by severe class imbalance, intricate attribute co‐dependencies, and domain shifts. We introduce VLM‐PAR, a modular vision–language framework built on frozen SigLIP 2 multilingual encoders. By first aligning image and prompt embeddings via refining visual features through a compact cross‐attention fusion, VLM‐PAR achieves interesting accuracy improvement on the highly imbalanced PA100K benchmark setting a new state of the art performance, while also delivering significant gains in mean accuracy across PETA and Market-1501 benchmarks. These results underscore the efficacy of integrating large-scale vision–language pretraining with targeted cross-modal refinement to overcome imbalance and generalization challenges in PAR.


\end{abstract}
\vspace{-12pt}
\section{Introduction}

\begin{figure}[t!]
    \centering
    \includegraphics[width=0.60\linewidth]{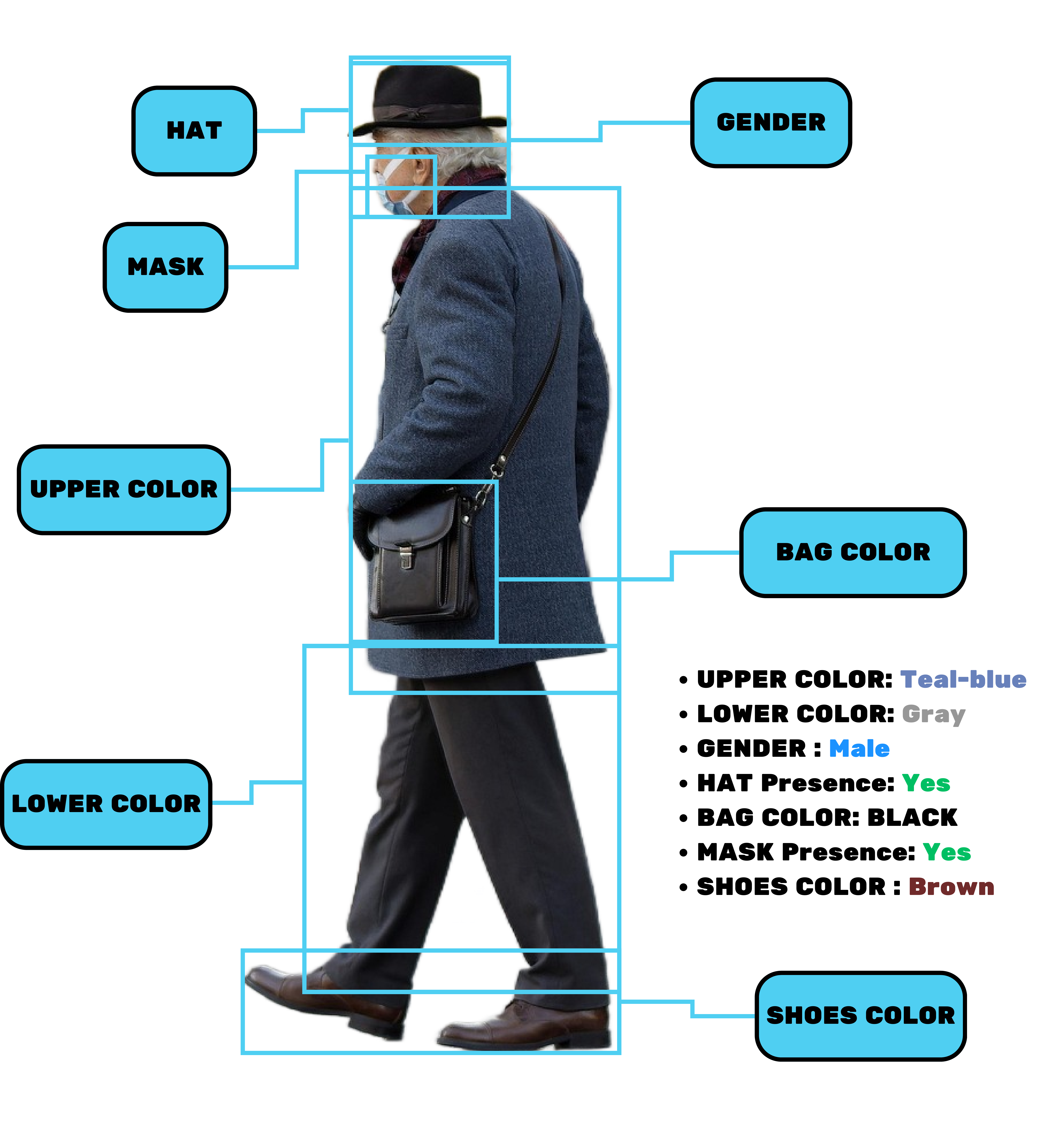}
    \caption{Example of pedestrian attribute annotations showing upper‐body clothing color, lower‐body clothing color, Mask, Shoes color, gender, bag color, and hat presence \cite{surprising_media_2021_pixabay}.}
    \label{fig:Par_example}
\end{figure}

Pedestrian Attribute Recognition (PAR) aims to automatically infer descriptive attributes from images of pedestrians, such as clothing color, gender, bag presence, and hat presence, enabling fine‐grained understanding in applications like surveillance, retail analytics, and autonomous driving (see Figure \ref{fig:Par_example}). Despite numerous innovations, existing PAR models struggle with severe class imbalance across attribute labels, limited cross‐domain generalization when deployed in novel environments, and the challenge of modeling inter‐attribute co‐dependence without negative transfer.
To address these challenges, we propose VLM-PAR, a modular vision–language framework built upon the SigLIP2 encoders \cite{tschannen2025siglip2}. In the first stage of our lightweight two‐stage fusion strategy, we freeze the pre-trained SigLIP 2 vision and text encoders and compute per‐attribute cosine similarity scores between image embeddings and their corresponding prompt embeddings. In the second stage, for each attribute, a compact multi‐head cross‐attention block refines the visual features by attending to the relevant text embedding. Task‐specific classification heads then map the fused features to discrete attribute predictions. This design preserves the generalization power of large‐scale pretraining while drastically reducing fine‐tuning overhead.
Our evaluations demonstrate that VLM-PAR not only achieves state‐of‐the‐art performance on standard PAR benchmarks such as PA-100K, PETA, and Market-1501—outperforming prior methods including AttriVision \cite{sedeh2025attrivision} and CLEAR \cite{bui2024clear} but also exhibits enhanced robustness to class imbalance and strong transferability across domains.

The rest of this paper is organized as follows. Section \ref{Relatedwork} reviews recent advances in pedestrian attribute recognition, particularly focusing on vision–language approaches such as AttriVision and CLEAR. In Section \ref{proposed}, we present the architecture of VLM-PAR in detail, including the SigLIP 2 encoders, the two‐stage fusion mechanism, and the design of task‐specific classification heads. Section \ref{setup} describes our experimental setup, including datasets, evaluation metrics, and implementation details. In Section \ref{results}, we report and analyze the obtained results. Finally, Section \ref{conclusion} concludes the paper and outlines directions for future work.

\section{Related Works} 
\label{Relatedwork}
Pedestrian Attribute Recognition (PAR) has long been formulated as a multi-label classification task, where each person's image is annotated with a set of semantic attributes. Early convolutional neural network (CNN) approaches extracted holistic or \emph{global} features from the image and trained per-attribute classifiers independently \cite{liu2017hydraplus}. While these methods were effective within controlled datasets, they often struggled with spatial misalignment and background noise. To address this, follow-up works introduced part-based modeling and attention mechanisms, enabling the network to focus on relevant body regions associated with specific attributes \cite{zhao2018grouping}. Despite their success on intra-dataset evaluations, these methods showed limited cross-domain generalization due to overfitting on dataset-specific cues.
Recent advances have explored transformer-based architectures and vision language paradigms to overcome the limitations of traditional CNNs. \emph{CLEAR} \cite{bui2024clear} proposed a cross-modal transformer incorporating pseudo-textual attribute descriptions and aligning them with image features through a cross-attention mechanism. While CLEAR achieves competitive performance, it incurs significant computational costs due to its complex multi-stream transformer backbone. Parallel to this, vision–language models (VLMs) such as CLIP \cite{radford2021learning} have emerged as powerful tools for zero-shot recognition. Leveraging CLIP's joint embedding space, \emph{AttriVision} \cite{sedeh2025attrivision} reformulates PAR as an image-text retrieval task, casting each attribute into a natural language prompt. By fine-tuning CLIP with a Focal Cross-Entropy loss tailored to class imbalance, AttriVision improves classification accuracy and generalization to unseen domains such as vehicle attribute recognition.
Building on this vision–language direction, Shen {\it et al.} \cite{shen2024sspnet} introduced \emph{SSPNet}, a Scale and Spatial Priors Guided Network that combines adaptive feature-scale selection with location priors to generate spatially interpretable and scale-aware representations. This framework improves robustness to variations in human pose and occlusion. Similarly, Li {\it et al.} \cite{li2025attribute} presented \emph{ACMFNet}, which enhances attribute reasoning through two novel modules: an Attribute-Correlation Query (ACQ) module that captures inter-attribute dependencies and a Mask-Fusion (MF) module that merges global and local representations, leading to more discriminative embeddings. Further refining transformer-based methods, Thakare {\it et al.} \cite{thakare2025clipping} proposed \emph{CLIP+CP}, which integrates contrastive prompt learning by dynamically generating attribute-specific prompts. This strategy enhances domain transferability by aligning attribute semantics across diverse visual domains, making it particularly effective in real-world, unconstrained settings.
Our work, \emph{VLM-PAR}, extends these approaches by leveraging \emph{SigLIP2} \cite{tschannen2025siglip2}, a state-of-the-art multilingual vision–language model. Rather than fine-tuning the entire VLM, we propose a lightweight two-stage fusion pipeline: (i) an initial cosine alignment step using frozen SigLIP2 image and text encoders to generate semantically meaningful embeddings, followed by (ii) dedicated multi-head cross-attention blocks that model attribute-specific dependencies. This modular architecture allows for efficient adaptation while preserving the generalization capabilities of large-scale VLMs. VLM-PAR achieves robust cross-domain performance with significantly reduced training overhead compared to prior prompt-based or transformer-only methods by freezing the backbone and optimizing only small fusion and classification heads.

\section{Proposed Method}
\label{proposed}
\subsection{Background knowledge}
\subsubsection{Siglip Model}
SigLIP 2 is a family of multilingual vision language encoders that builds directly on the original SigLIP training recipe \cite{Zhai2023SigLIP,tschannen2025siglip2}. In this second iteration, the training objective is extended with captioning‐based pretraining, self‐supervised losses, and online data curation, which yields significant gains in core capabilities such as zero‐shot classification, image–text retrieval, and transfer learning for downstream tasks. SigLIP 2 introduces two complementary architectural variants \emph{FixRes} and \emph{NaFlex} to accommodate diverse application requirements. The NaFlex variant extends the standard processing pipeline by ingesting images at multiple resolutions while rigorously preserving their native aspect ratios, yielding substantial gains on tasks where spatial fidelity is critical. To balance inference efficiency against representational capacity, four preconfigured scales (ViT‐B, ViT‐L, So400 M, and \emph{g}) are provided, enabling practitioners to select an optimal trade‐off for their computational budget. Empirical evaluations demonstrate that SigLIP 2 consistently surpasses its predecessor at every scale, delivering state‐of‐the‐art performance on multilingual classification benchmarks, dense prediction challenges, and precise localization tasks.
\vspace{-12pt}
\subsubsection{Cross-Attention}
Cross-attention is a mechanism where queries come from one sequence while keys and values derive from another. This allows the model to incorporate relevant context from multiple inputs effectively. It is a crucial component of encoder-decoder architectures in tasks like machine translation, where the decoder attends to encoder outputs to guide generation \cite{gheini-etal-2021-cross}.“Cross-Attention is All You Need” by Gheini {\it et al.} shows that fine-tuning only the cross-attention parameters during transfer learning is nearly as effective as updating the entire Transformer, underscoring the centrality of cross-attention in aligning representations across tasks \cite{gheini-etal-2021-cross}.
\vspace{-12pt} 
\subsection{VLM-PAR Architecture Overview}
VLM-PAR, illustrated in Figure \ref{fig:VLM-PAR}, is built on a modular paradigm in which each component is responsible for a specific facet of the person attribute recognition task. The architecture ingests an image and a set of textual prompts, each about one of five attributes. An efficient fusion of vision and language features is achieved through a two‐stage process: an initial semantic alignment via cosine similarity, followed by a more granular interaction in cross‐attention modules. The final step applies attribute-specific classification heads that produce probabilities over discrete categories. By freezing the heavy-weight SigLIP2 encoders, the model leverages extensive pretraining while reducing fine-tuning overhead; the lightweight fusion and classification layers then adapt these rich representations to the specific downstream attribute tasks.
\begin{figure}
    \centering
    \includegraphics[width=1.0\linewidth]{arch_vlm_1.png}
    \caption{VLM-PAR pipeline: a SigLIP-based image encoder and text encoder process person images and attribute-specific questions, align visual and textual features via multi-head cross-attention, then feed the attended features into task-specific classifiers for multi-attribute prediction.}
    \label{fig:VLM-PAR}
\end{figure}
\vspace{-10pt} 
\subsection{SigLIP2 Encoder Components}
\subsubsection{Vision Encoder (ViT-B/16)}
The vision encoder transforms the input image $I\in\mathbb{R}^{H\times W\times3}$ into a compact embedding $f_{\text{img}}\in\mathbb{R}^{768}$. To accomplish this, the image is divided into $16\times16$ patches indexed by $p\in\{1,\ldots,196\}$, each flattened to a vector of size $16\times16\times3$. A linear projection defined by $\mathbf{W}_E\in\mathbb{R}^{768\times768}$ maps each flattened patch into the model dimension, and positional embeddings $\mathbf{e}_{\text{pos}}^p$ are added to retain spatial configuration:
\begin{equation}
  \mathbf{x}_{\text{patch}}^p = \mathbf{W}_E\,\mathrm{vec}(I_p) + \mathbf{e}_{\text{pos}}^p.
\end{equation}
Subsequently, a series of 12 transformer layers applies multi-head self-attention (MSA) and feed-forward transforms. Within each layer, queries, keys, and values are derived from the preceding hidden states, enabling the model to learn contextual relations across all patches. This mechanism allows fine-grained features such as fabric texture, seam lines, or accessory shapes to influence the pooled output, and it supports global consistency by modeling long-range dependencies. The iterative update rule is given by:
\begin{align}
  \mathbf{h}_\ell^v &= \mathrm{LayerNorm}\bigl(\mathrm{MSA}(\mathbf{h}_{\ell-1}^v) + \mathbf{h}_{\ell-1}^v\bigr),\\
  \mathbf{z}_\ell^v &= \mathrm{LayerNorm}\bigl(\mathrm{MLP}(\mathbf{h}_\ell^v) + \mathbf{h}_\ell^v\bigr).
\end{align}
The final layer of the vision encoder outputs a matrix of patch embeddings:
\begin{equation}
\label{eq:patch_embeddings}
f_{\mathrm{img}} = \mathbf{z}_{12}^v[1\!:\!,:] \in \mathbb{R}^{N\times768},
\end{equation}
where \(\mathbf{z}_{12}^v\in\mathbb{R}^{(N+1)\times768}\) denotes the final-layer token embeddings including a leading [CLS] vector and slicing \(\mathbf{z}_{12}^v[1\!:\!,:]\) isolates the \(N\) spatial patch representations.
\subsubsection{Text Encoder}
Each attribute question $q_i$ (e.g., “What color is the hat?”) is first processed by the SigLip 2 tokenizer, yielding a sequence of $T$ tokens. These tokens are embedded via a matrix $\mathbf{E}_{\text{token}}\in\mathbb{R}^{32000\times768}$ (where \(32000\) is the vocabulary size) and enhanced with positional embeddings $\mathbf{e}_t^{\text{pos}}$, yielding
\begin{equation}
\mathbf{x}_{\text{token}}^{(i,t)} = \mathbf{E}_{\text{token}}(q_i^t) + \mathbf{e}_t^{\text{pos}}.
\end{equation}
The text encoder consists of 12 causal transformer layers that preserve the autoregressive structure. In each layer, the hidden states are updated according to:
\begin{align}
  \mathbf{h}_\ell^t &= \mathrm{LayerNorm}\bigl(\mathrm{CausalMSA}(\mathbf{h}_{\ell-1}^t) + \mathbf{h}_{\ell-1}^t\bigr),\\
  \mathbf{z}_\ell^t &= \mathrm{LayerNorm}\bigl(\mathrm{MLP}(\mathbf{h}_\ell^t) + \mathbf{h}_\ell^t\bigr).
\end{align}
Extracting the full sequence of token embeddings ensures compatibility with cross-attention operations:
\begin{equation}
  f_{\text{text}}^{(i)} = 
  \left[\mathbf{z}_{12}^{(i,1)}, \mathbf{z}_{12}^{(i,2)}, \dots, \mathbf{z}_{12}^{(i,T)}\right] 
  \in \mathbb{R}^{T \times 768},
\end{equation}
where $\mathbf{z}_{12}^{(i,t)} \in \mathbb{R}^{768}$ denotes the embedding of the $t$-th token in the $i$-th attribute prompt, and $T$ is the sequence length. This formulation retains all tokens to ensure proper alignment with image features during cross-attention.

\subsection{Cross-Attention Fusion}
Each attribute undergoes fusion through a dedicated multi-head cross-attention block that learns attribute-specific vision-language alignments. In this block, the image feature $f_{\text{img}} \in \mathbb{R}^{N \times 768}$ (where $N$ is the number of image patches) is projected to queries $Q_i$, while $f_{\text{text}}^{(i)} \in \mathbb{R}^{T \times 768}$ (from the text encoder) produces keys $K_i$ and values $V_i$, each via learnable matrices $W_Q^{(i)}, W_K^{(i)}, W_V^{(i)} \in \mathbb{R}^{768 \times 768}$:
\begin{equation}\label{eq:cross_attn_projections}
  Q_i = f_{\text{img}}W_Q^{(i)} \in \mathbb{R}^{N \times 768}, \quad
  K_i = f_{\text{text}}^{(i)}W_K^{(i)} \in \mathbb{R}^{T \times 768}, \quad
  V_i = f_{\text{text}}^{(i)}W_V^{(i)} \in \mathbb{R}^{T \times 768}.
\end{equation}
Each attribute $i$ maintains its projection matrices to enable specialized feature alignment. This design allows different attributes to focus on distinct visual regions and semantic concepts. For instance, The dedicated transformations prevent negative transfer between dissimilar attributes and enable each to learn optimal query-key-value mappings for its specific recognition task.
Each cross-attention block employs 8 attention heads with $d_k=96$ dimensions per head. The projection matrices are decomposed across heads as $W_Q^{(i)} = [W_{Q,1}^{(i)}, \ldots, W_{Q,8}^{(i)}]$ where each $W_{Q,h}^{(i)} \in \mathbb{R}^{768 \times 96}$. This multi-head design enables parallel processing of complementary attention patterns focusing on color, texture, spatial location, and other visual attributes.
Each head computes scaled dot-product attention:
\begin{equation}
\text{head}_h^{(i)}
=
\mathrm{softmax}\!\Bigl(
  \frac{1}{\sqrt{d_k}}
  Q_i^{(h)} \bigl(K_i^{(h)}\bigr)^\top
\Bigr)\,
V_i^{(h)}.
\end{equation}
Here, $Q_i$ queries spatial locations, $K_i$ encodes semantic content, and their similarity drives attention weights that refine visual embeddings. The concatenated outputs are projected and residually combined with $f_{\text{img}}$:
\begin{align}
  a_i &= \text{Concat}(\text{head}_1^{(i)},\dots,\text{head}_8^{(i)})W_O^{(i)},\\
  h_i &= \text{LayerNorm}(a_i + f_{\text{img}}).
\end{align}
This residual fusion preserves low-level visual details while incorporating high-level semantic cues. The resulting $h_i$ contains rich visual context and attribute-specific semantics optimized for classification.
\subsection{Multi-Task Classification Heads}
The fused embedding $h_i$ now contains both rich visual context and attribute-specific semantics. A lightweight linear layer then maps $h_i$ to logits $z_i$:
\begin{equation}
  z_i = h_i W_{\text{cls}}^{(i)} + b_{\text{cls}}^{(i)},
\end{equation}
followed by a softmax activation to produce a probability distribution $p_i$:
\begin{equation}
  p_i = \text{softmax}(z_i)
\end{equation}
Each attribute $i$ employs a dedicated classification head with parameters $W_{\text{cls}}^{(i)}$ and $b_{\text{cls}}^{(i)}$ tailored to its output space. Categorical attributes project to $K_i$ dimensional space while binary attributes output 2-dimensional logits.
All classification heads are trained jointly in a single model, processing $N$ attributes simultaneously for efficient computation and knowledge sharing. The total loss combines individual attribute losses:
 
\begin{equation}
  \mathcal{L}_{\text{total}} = \frac{1}{N} \sum_{i=1}^{N} \mathcal{L}_i(p_i, y_i^{\text{true}})
\end{equation}
where each $\mathcal{L}_i$ incorporates cross-entropy, focal loss, and label smoothing components.
Task-specific heads mitigate negative transfer while enabling knowledge sharing through joint optimization. Each attribute benefits from tailored parameters and shared visual representations.
The final model output for each attribute $i$
\begin{equation}
  \text{Output}_i = \{\hat{y}_i, c_i\}
\end{equation}
where $\hat{y}_i = \arg\max(p_i)$ is the predicted class and $c_i = \max(p_i)$ represents the prediction confidence.

\section{Experimental Setup}
\label{setup}
\subsection{Datasets}
In this work, we use three widely adopted datasets for PAR: PA-100K \cite{liu2017hydraplus}, PETA \cite{deng2014pedestrian}, and Market1501 \cite{zheng2015scalable,lin2019improving}.
PA-100K is a large-scale dataset specifically designed for PAR tasks. It contains 100,000 images of pedestrians captured from real-world surveillance scenarios, each annotated with 26 binary attributes covering clothing types, accessories, and human actions. The dataset is divided into 80,000 training images, 10,000 for validation, and 10,000 for testing, enabling standardized and reproducible benchmarking.

PETA consists of 19,000 pedestrian images collected from various datasets and is annotated with a rich set of 65 attributes, including 61 binary and 4 multi-class labels that cover aspects such as gender, age group, clothing style, and accessories. Following standard practice, we use 9,500 images for training, 1,900 for validation, and 7,600 for testing.

Market-1501 was initially introduced for person re-identification but has since been extended for pedestrian attribute recognition (PAR) tasks. It contains over 30,000 images of 1,501 identities captured from six camera views. The dataset has been annotated with 27 binary attributes covering gender, age, body type, clothing, and accessories, making it suitable for PAR-related experiments.

\begin{figure}[h]
    \centering
    \includegraphics[width=1.0\linewidth]{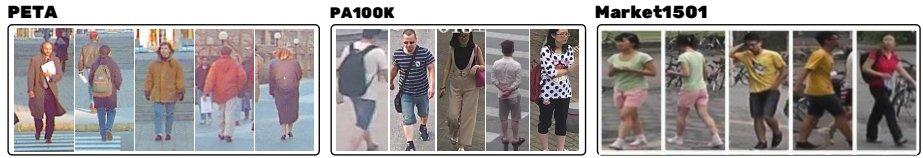}
    \vspace{-10pt}
    \caption{Examples of Pedestrian images from three widely-used attribute datasets: PETA, PA‑100K, and Market‑1501.}
    \label{fig:datasets}
\end{figure}
 
\vspace{-30pt} 
\subsection{Metrics}
We evaluate all methods using mean accuracy (mA) and F1-score. Mean accuracy averages per-attribute recall on positive and negative samples, and F1-score is the harmonic mean of precision and recall:
\[
\text{mA} = \frac{1}{2A} \sum_{i=1}^{A} \left(\frac{TP_i}{TP_i + FN_i} + \frac{TN_i}{TN_i + FP_i}\right), \quad \text{F1}_i = \frac{2 \cdot \text{Precision}_i \times \text{Recall}_i}{\text{Precision}_i + \text{Recall}_i},
\]
where $A$ is the total number of attributes and $TP_i$, $TN_i$, $FP_i$, $FN_i$ denote true/false positives/negatives for attribute $i$. Overall F1 is averaged across all attributes. Higher values indicate better performance.

\section{Experimental Results and Discussion}
\label{results}

Table~\ref{tab:results} reports the detailed comparison of VLM-PAR against prior methods across four benchmarks. On the PA100K dataset, VLM-PAR achieves a 3.08\% higher mean accuracy than AttriVision, indicating substantially better overall coverage of positive and negative samples. Its F1, however, drops by 0.78\%  reflecting a recall‑oriented fusion that better captures rare attributes at the expense of a few extra false positives. 

On PETA, we observe a 2.82\% jump in mean accuracy and a pronounced 1.14\% uplift in F1. This significant F1 gain implies that VLM-PAR is particularly effective at handling the extreme class imbalance present in PETA, correctly identifying rare attributes while maintaining low error rates on common ones. The cross-attention fusion sharpens attribute-specific cues, such as subtle clothing details, that other models miss.

For Market-1501, the mean accuracy of VLM-PAR increases by 1.58\%, confirming consistent gains even in datasets with simpler attribute distributions. However, the 9.63\%  drop in F1 highlights a shift in the precision-recall trade-off. Our model tends to make more positive predictions, improving raw coverage (accuracy) but at the cost of additional false positives. This suggests that further calibration, either via more conservative classification thresholds or specialized loss weighting, could recover F1 without sacrificing accuracy.
\renewcommand{\arraystretch}{1.0} 
\setlength{\tabcolsep}{4pt}      

\begin{table}[ht]
\centering
\caption{Comparison of mean Accuracy (mA) and F1 scores on standard PAR benchmarks. Highest scores in bold, second-highest underlined.}
\label{tab:results}
\begin{tabular}{l cc cc cc c}
\toprule
\textbf{Method} 
  & \multicolumn{2}{c}{\textbf{PA100K}} 
  & \multicolumn{2}{c}{\textbf{PETA}} 
  & \multicolumn{2}{c}{\textbf{Market1501}} \\
 & mA & F1 & mA & F1 & mA & F1 \\
\midrule
SSPNet \cite{shen2024sspnet} 
  & 83.58  & 88.55  & 88.73  & 89.50  & --      & --     \\ 
ACMFNet \cite{li2025attribute} 
  & 83.71  & 87.74  & 87.52  & 87.31  & --      & --  \\ 
CLEAR \cite{bui2024clear} 
  & 87.20  & 91.00  & 88.20  & 89.80  & 83.00  & \underline{87.90}        \\ 
CLIP+CP \cite{thakare2025clipping} 
  & 86.45   & \underline{92.77}  & 90.05  & \underline{91.86}  & --      & --           \\ 
AttriVision \cite{sedeh2025attrivision} 
  & \underline{89.80}  & \textbf{93.10}  & \underline{90.70}  & 91.50  & \underline{83.80}  & \textbf{88.80}       \\ 
\textbf{VLM-PAR (Ours)} 
  & \textbf{92.88}  & 92.32  & \textbf{93.52}  & \textbf{92.64}  & \textbf{85.38}  & 79.17  \\ 
\bottomrule
\end{tabular}
\end{table}

\begin{figure}
    \centering
    \includegraphics[width=1.0\linewidth]{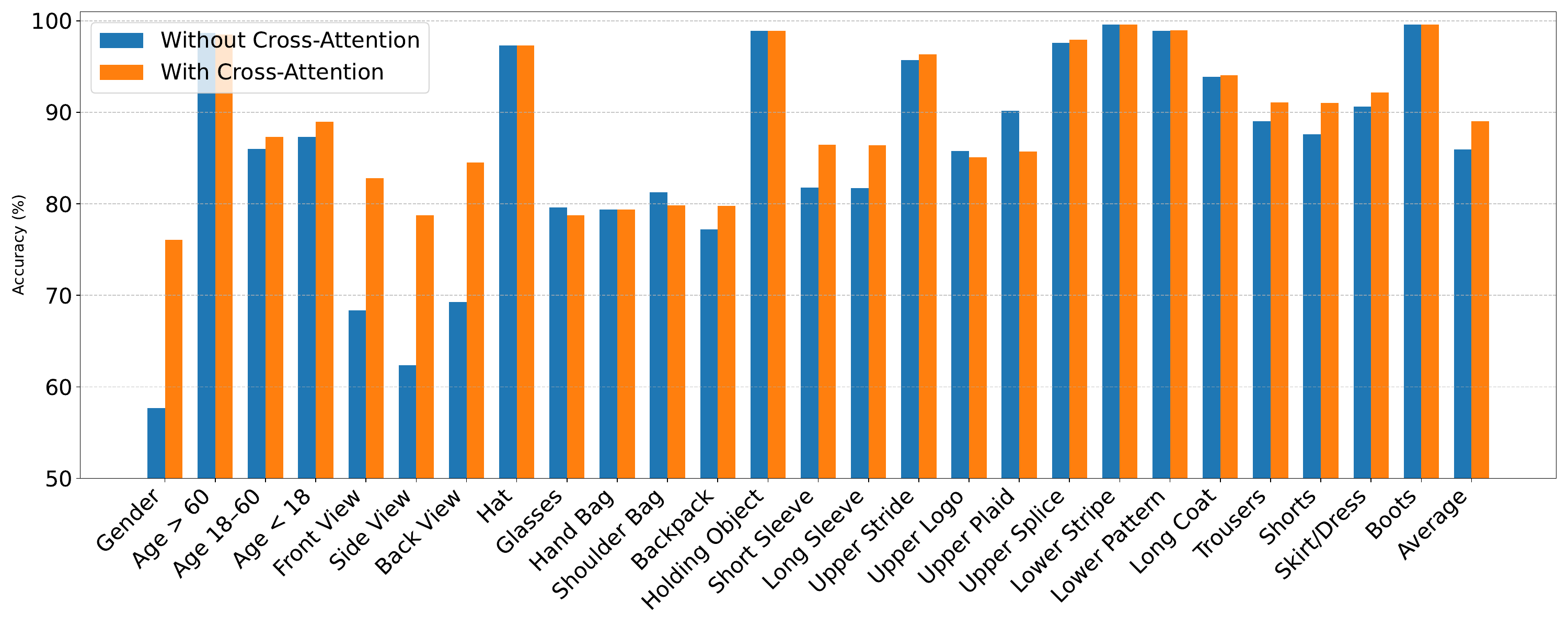}
    \caption{Ablation study comparing attribute classification accuracy with and without cross-attention.}
    \label{fig:ablation_study}
\end{figure}

\renewcommand{\arraystretch}{1.0} 
To evaluate the impact of the proposed cross-attention mechanism in VLM-PAR, we performed an ablation study using the PA-100K dataset. The model was trained for one epoch, and the validation accuracy was measured with and without the cross-attention module (text part is not used; only image part in this case). Figure~\ref{fig:ablation_study} shows a comparison of classification accuracy for 26 pedestrian attributes.
The results show a consistent improvement when cross-attention is used. For example, the accuracy for the attribute \textit{Front View} increased significantly from 68.36\% to 82.81\%, while \textit{Side View} improved from 62.34\% to 78.75\% and \textit{Back View} from 69.30\% to 84.53\%. This shows that cross-attention improves the model's ability to understand pose and orientation.
For appearance-related attributes, we observe meaningful improvements as well. The attribute \textit{Short Sleeve} increased from 81.80\% to 86.48\%, and \textit{Trousers} from 90.62\% to 92.19\%. For some attributes such as \textit{Skirt/Dress} there was no change (they remained at 99.61\%), suggesting that these attributes may already be well learnt without the need for additional attention mechanisms. For a few attributes such as \textit{Upper Logo}, a slight decrease was observed (from 85.78\% to 85.08\%), but this is relatively insignificant.
Overall, the average classification accuracy improved from 85.98\% without cross-attention to 89.06\% with it. These results confirm that cross-attention plays an important role in improving attribute recognition, especially for more complex or view-dependent attributes.

\section{Conclusion}
\label{conclusion}
In this work, we introduced VLM-PAR, a novel vision–language framework for pedestrian attribute recognition that leverages the powerful SigLIP 2 encoders alongside a lightweight two-stage fusion mechanism. By freezing the pre-trained image and text backbones and applying attribute-specific multi-head cross-attention, VLM-PAR achieves substantial gains in mean accuracy and F1 across multiple standard benchmarks PA100K, PETA, and Market-1501—while maintaining minimal fine-tuning overhead. Our experiments demonstrate that the fusion of generic multilingual vision–language representations with targeted cross-modal refinement can simultaneously address class imbalance, cross-domain generalization, and inter-attribute co-dependence.
Looking ahead, we plan to explore adaptive thresholding and dynamic loss weighting to better calibrate the precision-recall trade-off, particularly in scenarios where F1 performance lags. We also aim to investigate prompt engineering and text augmentation strategies to enrich linguistic descriptions and enhance zero-shot transfer to unseen attributes. 


\vspace{-12pt}
\bibliography{references}

@inproceedings{deng2014pedestrian,
  title={Pedestrian attribute recognition at far distance},
  author={Deng, Yubin and Luo, Ping and Loy, Chen Change and Tang, Xiaoou},
  booktitle={Proceedings of the 22nd ACM international conference on Multimedia},
  pages={789--792},
  year={2014}
}

@inproceedings{zheng2015scalable,
  title={Scalable person re-identification: A benchmark},
  author={Zheng, Liang and Shen, Liyue and Tian, Lu and Wang, Shengjin and Wang, Jingdong and Tian, Qi},
  booktitle={Proceedings of the IEEE International Conference on Computer Vision},
  year={2015}
}

@inproceedings{liu2017hydraplus,
  title={Hydraplus-net: Attentive deep features for pedestrian analysis},
  author={Liu, Xihui and Zhao, Haiyu and Tian, Maoqing and Sheng, Lu and Shao, Jing and Yi, Shuai and Yan, Junjie and Wang, Xiaogang},
  booktitle={Proceedings of the IEEE international conference on computer vision},
  pages={350--359},
  year={2017}
}

@article{lin2019improving,
  title={Improving Person Re-identification by Attribute and Identity Learning},
  author={Lin, Yutian and Zheng, Liang and Zheng, Zhedong and Wu, Yu and Hu, Zhilan and Yan, Chenggang and Yang, Yi},
  journal={Pattern Recognition},
  doi = {https://doi.org/10.1016/j.patcog.2019.06.006},
  year={2019}
}

@article{bui2024clear,
  title={Clear: Cross-transformers with pre-trained language model is all you need for person attribute recognition and retrieval},
  author={Bui, Doanh C and Le, Thinh V and Ngo, Ba Hung and Choi, Tae Jong},
  journal={arXiv preprint arXiv:2403.06119},
  year={2024}
}

@article{shen2024sspnet,
  title={SSPNet: Scale and spatial priors guided generalizable and interpretable pedestrian attribute recognition},
  author={Shen, Jifeng and Guo, Teng and Zuo, Xin and Fan, Heng and Yang, Wankou},
  journal={Pattern Recognition},
  volume={148},
  pages={110194},
  year={2024},
  publisher={Elsevier}
}

@inproceedings{sedeh2025attrivision,
  title={AttriVision: Advancing Generalization in Pedestrian Attribute Recognition using CLIP},
  author={Sedeh, Mehran Adibi and Benbihi, Assia and Martin, Romain and Clausel, Marianne and Pradalier, C{\'e}dric},
  booktitle={Proceedings of the Winter Conference on Applications of Computer Vision},
  pages={354--365},
  year={2025}
}

@inproceedings{thakare2025clipping,
  title={CLIPping Imbalances: A Novel Evaluation Baseline and PEARL Dataset for Pedestrian Attribute Recognition},
  author={Thakare, Kamalakar Vijay and Lohani, Lalit and Nayak, Kamakshya Prasad and Dogra, Debi Prosad and Choi, Heeseung and Jung, Hyungjoo and Kim, Ig-Jae},
  booktitle={2025 IEEE/CVF Winter Conference on Applications of Computer Vision (WACV)},
  pages={7102--7111},
  year={2025},
  organization={IEEE}
}

@article{li2025attribute,
  title={Attribute correlation mask fusion network for pedestrian attribute recognition},
  author={Li, Baoan and Zhang, Long and Teng, Shangzhi and Lyu, Xueqiang},
  journal={The Visual Computer},
  volume={41},
  number={6},
  pages={3719--3734},
  year={2025},
  publisher={Springer}
}

@article{Tschannen2025SigLIP2,
  title        = {SigLIP 2: Multilingual Vision–Language Encoders with Improved Semantic Understanding, Localization, and Dense Features},
  author       = {Tschannen, Michael and Gritsenko, Alexey and Wang, Xiao and Naeem, Muhammad Ferjad and Alabdulmohsin, Ibrahim and Parthasarathy, Nikhil and Evans, Talfan and Beyer, Lucas and Xia, Ye and Mustafa, Basil and Hénaff, Olivier and Harmsen, Jeremiah and Steiner, Andreas and Zhai, Xiaohua},
  journal      = {arXiv preprint arXiv:2502.14786},
  year         = {2025},
  url          = {https://arxiv.org/abs/2502.14786}
}

@inproceedings{Zhai2023SigLIP,
  title        = {{Sigmoid Loss for Language Image Pre-Training (SigLIP)}},
  author       = {Zhai, Xiaohua and Mustafa, Basil and Kolesnikov, Alexander and Beyer, Lucas},
  booktitle    = {arXiv preprint arXiv:2303.15343},
  year         = {2023},
  url          = {https://arxiv.org/abs/2303.15343}
}

@inproceedings{zhao2018grouping,
  title        = {Grouping Attribute Recognition for Pedestrian with Joint Recurrent Learning},
  author       = {Zhao, Xin and Sang, Liufang and Ding, Guiguang and Guo, Yuchen and Jin, Xiaoming},
  booktitle    = {Proceedings of the European Conference on Computer Vision Workshops (ECCVW)},
  year         = {2018},
  publisher    = {Springer},
  note         = {pp. 0--0}
}

@inproceedings{gheini-etal-2021-cross,
  title     = {Cross-Attention is All You Need: Adapting Pretrained Transformers for Machine Translation},
  author    = {Gheini, Mozhdeh and Ren, Xiang and May, Jonathan},
  booktitle = {Proceedings of the 2021 Conference on Empirical Methods in Natural Language Processing (EMNLP)},
  year      = {2021},
  pages     = {1754--1765},
  publisher = {Association for Computational Linguistics},
  doi       = {10.18653/v1/2021.emnlp-main.132}
}

@misc{surprising_media_2021_pixabay,
  author       = {{Surprising\_Media}},
  title        = {{Elderly man, Crossing the street}},
  howpublished = {\url{https://pixabay.com/photos/elderly-man-crossing-the-street-6792215/}},
  year         = {2021},
  month        = nov # "~14",
  note         = {Photograph. Accessed 2025-06-12},
}

@inproceedings{radford2021learning,
  title        = {Learning transferable visual models from natural language supervision},
  author       = {Radford, Alec and Kim, Jong Wook and Hallacy, Chris and Ramesh, Aditya and Goh, Gabriel and Agarwal, Sandhini and Sastry, Girish and Askell, Amanda and Mishkin, Pamela and Clark, Jack and et al.},
  booktitle    = {International Conference on Machine Learning},
  pages        = {8748--8763},
  year         = {2021},
  organization = {PMLR}
}
\bibliographystyle{splncs04}
\end{document}